\documentclass[10pt,twocolumn,letterpaper]{article}
\PassOptionsToPackage{table}{xcolor}
\usepackage[pagenumbers]{cvpr}
\usepackage{multirow}
\usepackage{microtype}
\usepackage{tikz}
\graphicspath{{figures/}}
\definecolor{cvprblue}{rgb}{0.21,0.49,0.74}
\usepackage[pagebackref,breaklinks,colorlinks,allcolors=cvprblue]{hyperref}
\title{How Well Do Segmentation Foundation Models Generalize Across Structural and Functional Medical Imaging?}
\author{%
Yichi Zhang\textsuperscript{1,2}\quad
Le Xue\textsuperscript{1}\quad
Feiyang Xiao\textsuperscript{1}\quad
Wenbo Zhang\textsuperscript{1,2}\quad
Gang Feng\textsuperscript{3}\\
Chenguang Zheng\textsuperscript{1,2}\quad
Yuan Qi\textsuperscript{1,2}\quad
Yuan Cheng\textsuperscript{1,2}\quad
Zixin Hu\textsuperscript{1,2}\\[4pt]
{\normalsize \textsuperscript{1}Fudan University\quad
\textsuperscript{2}Shanghai Academy of Artificial Intelligence for Science}\\
{\normalsize \textsuperscript{3}Shanghai Universal Medical Imaging Diagnostic Center}
}

\hypersetup{
  pdftitle={How Well Do Segmentation Foundation Models Generalize Across Structural and Functional Medical Imaging?},
  pdfauthor={Yichi Zhang, Le Xue, Feiyang Xiao, Wenbo Zhang, Gang Feng, Chenguang Zheng, Yuan Qi, Yuan Cheng, Zixin Hu}
}
\begin{document}
\maketitle
\begin{abstract}
Segmentation foundation models offer reusable tools for universal medical-image segmentation, but their development and evaluation have largely centered on structural CT and MRI imaging. Functional imaging such as PET provides complementary metabolic information yet remains less studied and validated. Moreover, existing evaluations often involve different datasets and tasks, making modality-associated variation difficult to distinguish from differences in segmentation targets and data distributions. To enable a more direct comparison, we introduce UMD, a whole-body multi-organ dataset comprising 490 PET/CT and 464 PET/MRI examinations with 13 annotated organs (approximately 675,000 2D slices and 12,000 3D organ annotations). UMD evaluates structural and functional images separately on corresponding organs within the same examination, reducing variation in patient anatomy and target selection. We benchmark the out-of-the-box segmentation capabilities of five representative 3D foundation models on UMD.
Overall, most evaluated models show limited PET segmentation performance, and some configurations also yield near-zero overlap on CT or MRI under the evaluated workflows.
These findings highlight the need to assess tool applicability across different imaging conditions and identify opportunities for cross-modality generalization. UMD provides a paired data resource and evaluation setting for this investigation. The dataset will be public at https://github.com/YichiZhang98/UMD.
\end{abstract}

\begin{figure}[t]
\centering
\input{figures/figure1_training}
\caption{Training-data composition of representative segmentation foundation models. The composition of imaging modalities reveals a structural bias.   The distribution of anatomical regions demonstrates a strong bias toward the abdomen.   Conversely, data representing the pelvis and skeleton are persistently scarce.   Together, these distributions highlight systemic data imbalances that may limit the generalization of current models to comprehensive, whole-body, and multi-modal tasks.}
\label{datafig}
\end{figure}

\begin{figure*}[t]
\centering
\input{figures/figure2_overview}
\caption{UMD dataset and benchmark design. (a) Comparisons across different datasets often vary in cohort, anatomical target, and modality. (b) UMD enables comparisons of corresponding organ targets in paired PET/CT and PET/MRI examinations. Pairing reduces differences in patient anatomy and target selection, while acquisition, organ visibility, reference-label construction, and inference procedures remain relevant. (c) Acquisition distributions of the two cohorts.}
\label{overview}
\end{figure*}

\section{Introduction}
Medical-image segmentation supports quantitative measurements of organ volumes, disease burden, and treatment response~\cite{shen2017deep,ilesanmi2024reviewing}. Deep learning models have shown strong potential in medical image segmentation by extracting intricate features from massive annotated datasets and achieving accurate segmentation across a diverse range of tasks ~\cite{antonelli2022medical,AbdomenCT-1K,lalande2021deep,gatidis2024results}.
Despite these advances, a prominent bottleneck of existing models remains their strict task-specificity. These models are typically designed and trained for specific predefined tasks and often suffer from severe performance degradation when exposed to new tasks or out-of-distribution data. Consequently, their capacity to generalize to other clinical scenarios is highly constrained, necessitating the development of distinct models for each specific imaging modality or target organ, which poses a major obstacle to broad clinical translation given that routine clinical workflows demand the robust segmentation of highly heterogeneous structures across a wide array of imaging modalities \cite{ma2023towards}.

Recent advancement of artificial intelligence has been defined by the emergence of foundation models (FMs), which are large-scale, pre-trained models capable of performing multi-task segmentation within a unified framework \cite{awais2025foundation,zhang2024challenges}. 
The field of medical image analysis is undergoing a paradigm shift from task-specific models to general-purpose foundation models.
Inspired by the success of the Segment Anything Model (SAM) in computer vision \cite{SAM}, the community has witnessed a proliferation of efforts to adapt this paradigm to medical image segmentation for quantitative image analysis \cite{mazurowski2023segment,zhang2024segment}.
While pretrained foundation models offer reusable representations and prompt-based interfaces that can reduce the need to develop separate models for each target, understanding when these tools can be used without task-specific fine-tuning is an important question for their practical adoption.

In radiology practice, structural and functional imaging provide a useful setting for this distinction. Structural imaging like Computed Tomography (CT) and Magnetic Resonance Imaging (MRI) depict anatomical structure through different physical signals, whereas functional imaging like Positron Emission Tomography (PET) depicts radiotracer uptake. These modalities differ in spatial resolution, contrast, and organ visibility. 
While these segmentation foundation models claim to provide a unified solution for diverse clinical tasks, their validation remains largely restricted to structural imaging (Fig.~\ref{datafig}). This leaves a significant modality discrepancy unexplored, particularly the divergence between anatomical structure and functional metabolism. Besides, most existing evaluations rely on test splits from aggregated datasets, overlooking performance under realistic out-of-distribution scenarios.

Existing benchmarks address complementary dimensions of segmentation performance. The Medical Segmentation Decathlon~\cite{antonelli2022medical} spans ten tasks with CT and MRI datasets, supporting assessment across different anatomical and pathological targets. Touchstone~\cite{bassi2024touchstone} emphasizes abdominal CT segmentation across institutions and evaluation distributions. These resources establish valuable settings for task diversity, external evaluation, and lesion analysis.
However, conventional cross-dataset evaluations can also obscure cross-modality robustness because different modalities are often paired with different cohorts, anatomical targets, and annotation protocols. Under this setup, a performance difference cannot be attributed to modality alone: task difficulty, organ morphology, acquisition, and imaging contrast vary simultaneously, as shown in Figure \ref{overview}(a).
This motivates a paired benchmark that controls patient anatomy and target definitions while preserving the model-specific inference process used in practice.

To bridge this gap, we introduce \textbf{UMD}, a benchmark of 490 paired PET/CT and 464 paired PET/MRI for whole-body multi-organ segmentation (Fig.~\ref{overview}(b)).
UMD contributes a new structural--functional imaging resource and a paired evaluation setting for released 3D foundation models as out-of-the-box segmentation tools. Each modality is supplied separately to the evaluated model, enabling comparisons on corresponding targets across imaging conditions.
This paired design, with multi-modal scans acquired within the same diagnostic examination from the same patient cohort, ensures that segmentation results for identical organs can be used to objectively compare performance differences across modalities.
The paired design reduces variation in patient anatomy and target definition.
The imaging data were newly collected rather than aggregated from existing datasets, which eliminates the risk of data leakage to guarantee the validity and authenticity of the evaluation results.

Our evaluation addresses three related questions: how out-of-the-box
performance varies across paired imaging modalities, which organ-level
capabilities are shared or differ across imaging conditions, and whether
the observed modality differences recur across individual examinations.
We evaluate five representative general-purpose segmentation foundation models through their specified prompting designs.
Notably, each of these frameworks has claimed to achieve superior zero-shot generalization performance across a wide array of 3D medical imaging tasks and anatomical structures.
Our evaluation shows that current segmentation foundation models perform differently across imaging modalities, with limited PET segmentation performance for most evaluated models.  UMD provides a new paired dataset and benchmark for assessing these capabilities across structural and functional imaging and supporting further research on cross-modality generalization.

\begin{table*}[t]
    \centering
    \small
    \setlength\tabcolsep{6pt}
    \renewcommand\arraystretch{1.3}
    \resizebox{\textwidth}{!}{%
\begin{tabular}{cc|ccc|cc|cc}
        \hline\hline
        \multirow{2}{*}{\textbf{Model}} & \multirow{2}{*}{\textbf{Setting}} & \multicolumn{3}{c|}{\textbf{Reported Performance$^{*}$ }} & \multicolumn{4}{c}{\textbf{UMD Evaluation Performance}} \\
        \cline{3-9} && CT & MRI & PET & CT & PET & MRI & PET \\
        \hline
        SAM-Med3D-turbo & 1 point prompt & 0.790 & 0.754 & / & 0.404 & 0.410 & 0.433 & 0.397 \\
        SegVol & 1 point prompt & 0.793 & / & / & 0.107 & 0.075 & 0.134 & 0.148 \\
        nnInteractive & 1 point prompt & $\sim$0.55 & $\sim$0.55 & / & 0.481 & 0.342 & 0.511 & 0.217 \\
        VISTA3D & class-id prompt & 0.711 & / & / & 0.838 & 0.010 & 0.022 & 0.008 \\
        SAT & text prompt & 0.756 & 0.838 & 0.634 & 0.018 & 0.003 & 0.006 & 0.000 \\ \hline
        nnUNet & 10 Training Samples & - & - & - & 0.686 & 0.605 & 0.652 & 0.560 \\
        \hline \hline
    \end{tabular}
}
    \caption{Overview of the performance of segmentation foundation models on their reported test sets and proposed UMD benchmark. Literature-reported scores ($*$) provide context for each source study and are not directly comparable with UMD because target organs and data distributions are different. $/$ denotes that the test data do not contain corresponding modality. $\sim$ denotes that the specific performance for each individual modality is not reported, and only the average performance is presented.}
    \label{result_performance}
\end{table*}

\section{Benchmark Design}

\subsection{Imaging background}

Structural and functional imaging provide complementary but physically different representations of the body. CT measures X-ray attenuation and expresses reconstructed values on the approximately standardized Hounsfield-unit scale. It generally provides high-resolution structural information, with strong contrast at interfaces such as air--lung, soft tissue--fat, and soft tissue--bone. MRI derives image contrast from nuclear-magnetic-resonance signals whose appearance depends on the pulse sequence, scanner, and acquisition parameters. MRI therefore provides rich soft-tissue contrast, but unlike CT, its intensity values do not have a universal quantitative scale across sequences and sites. PET is a functional imaging technique: after administration of a radiotracer, it estimates the spatial distribution of tracer uptake and thereby visualizes metabolic or molecular processes. PET commonly has lower spatial resolution and less sharply defined organ boundaries than CT or MRI, and normal physiological uptake can be intense in structures such as the brain, heart, and urinary bladder.

\subsection{Data acquisition and annotation}

UMD comprises 490 PET/CT and 464 PET/MRI examinations. Within each examination, structural and functional images were acquired from the same patient during a single diagnostic session and co-registered. Each examination corresponds to one patient.
Patients fasted for at least 6 hours with blood glucose levels confirmed at $< 11.1$ mmol/L prior to receiving an intravenous injection of $^{18}$F-FDG.
The scanning field was from the mid-thigh level to the top of the head with the subject in the supine position.
Voxel-wise annotations were provided for 13 diverse organs, including the liver, left kidney, right kidney, brain, heart, spleen, aorta, lung, colon, urinary bladder, pancreas, esophagus, and stomach.
Reference annotations were constructed using a two-stage pipeline.
Initial coarse segmentation masks were automatically generated via pre-trained segmentation tools \cite{totalsegmentator,totalsegmentator-mri}.
PET reference masks were generated by mapping the corresponding CT or MRI annotations to PET and then manually correcting the transferred masks. These annotations define anatomical organ targets informed by paired structural images.
All preliminary annotations were subsequently reviewed, corrected, and refined by radiologists using the LIFEx \cite{nioche2018lifex} platform.

\begin{table*}[t]
	\centering
    \small
    \setlength\tabcolsep{6pt}
	\renewcommand\arraystretch{1.1}
    \resizebox{\textwidth}{!}{%
\begin{tabular}{c|ccccc}
		\hline 	\hline
Model & SAM-Med3D-turbo & SegVol & nnInteractive & VISTA3D & SAT  \\ \hline
\multirow{2}{*}{Liver}
& \cellcolor{gray!15} \color{blue} 0.7039(0.1155) & \color{blue} 0.0806(0.1982) & \color{blue} \cellcolor{gray!40} 0.4993(0.3319) & \color{blue} \cellcolor{gray!40} 0.9186(0.0378) & \color{blue} 0.0057(0.0205) \\
& \color{red} 0.6835(0.1100) & \color{red} \cellcolor{gray!40} 0.1618(0.2539) & \color{red} 0.3904(0.3769) & \color{red} 0.0676(0.0475) & \color{red} 0.0020(0.0451)  \\ \hline
\multirow{2}{*}{Kidney-L}
& \color{blue} 0.4701(0.1300) & \color{blue} \cellcolor{gray!40} 0.1203(0.1967) & \color{blue} \cellcolor{gray!40} 0.7930(0.1636) & \color{blue} \cellcolor{gray!40} 0.8648(0.1007) & \color{blue} 0.0001(0.0018)  \\
& \color{red} \cellcolor{gray!40} 0.5678(0.1411) & \color{red} 0.0687(0.1329) & \color{red} 0.4464(0.3436) & \color{red} 0.0000(0.0000) & \color{red} 0.0041(0.0638)  \\ \hline
\multirow{2}{*}{Kidney-R}
& \color{blue} 0.4387(0.1491) & \color{blue} \cellcolor{gray!40} 0.1437(0.2279) &  \color{blue} \cellcolor{gray!40} 0.7780(0.1753) & \color{blue} \cellcolor{gray!40} 0.8365(0.1184) & \color{blue} 0.0061(0.0780) \\
& \color{red} \cellcolor{gray!40} 0.5596(0.1843) & \color{red} 0.0630(0.1178) & \color{red} 0.4187(0.3413) & \color{red} 0.0000(0.0000) & \color{red} 0.0082(0.0900) \\ \hline
\multirow{2}{*}{Brain}
& \color{blue} 0.6760(0.2378) & \color{blue} \cellcolor{gray!40} 0.3024(0.3926) & \color{blue} 0.7940(0.1432) & \color{blue} \cellcolor{gray!40} 0.9416(0.0468) & \color{blue} 0.0020(0.0451) \\
& \color{red} 0.6609(0.3030) & \color{red} 0.1399(0.2408) & \color{red} \cellcolor{gray!40} 0.8796(0.1250) & \color{red} 0.0478(0.0423) & \color{red} 0.0020(0.0451) \\ \hline
\multirow{2}{*}{Heart}
& \color{blue} \cellcolor{gray!40} 0.5073(0.0992) & \color{blue} 0.0993(0.1863) & \color{blue} \cellcolor{gray!40} 0.5607(0.2908) & \color{blue} \cellcolor{gray!40} 0.8018(0.0687) & \color{blue} 0.0020(0.0451) \\
& \color{red} 0.4581(0.1323) & \color{red} 0.1148(0.1994) & \color{red} 0.2214(0.2641) & \color{red} 0.0041(0.0181) & \color{red} 0.0041(0.0638) \\ \hline
\multirow{2}{*}{Spleen}
& \color{blue} 0.5110(0.1483) & \color{blue} 0.1208(0.2065) & \color{blue} \cellcolor{gray!40} 0.6064(0.2937) & \color{blue} \cellcolor{gray!40} 0.8595(0.0997) &  \color{blue} 0.0061(0.0780) \\
& \color{red} \cellcolor{gray!40} 0.5827(0.1295) &  \color{red} 0.1105(0.2000) & \color{red} 0.4378(0.3516) & \color{red} 0.0023(0.0154) &  \color{red} 0.0061(0.0780) \\ \hline
\multirow{2}{*}{Aorta}
& \color{blue} 0.1343(0.0588) & \color{blue} 0.0465(0.0900) & \color{blue} \cellcolor{gray!40} 0.2356(0.1325) & \color{blue} \cellcolor{gray!40} 0.8279(0.0496) & \color{blue} 0.0024(0.0452)  \\
& \color{red} \cellcolor{gray!40} 0.1635(0.0600) & \color{red} 0.0541(0.0894) & \color{red} 0.0662(0.0936) & \color{red} 0.0004(0.0049) &  \color{red} 0.0020(0.0451)  \\ \hline
\multirow{2}{*}{Lung}
& \color{blue} \cellcolor{gray!40} 0.6243(0.0767) & \color{blue} \cellcolor{gray!40} 0.1426(0.2239) & \color{blue} \cellcolor{gray!40} 0.4805(0.2361) & \color{blue} \cellcolor{gray!40} 0.9200(0.0509) & \color{blue} \cellcolor{gray!40} 0.0429(0.0151)  \\
& \color{red} 0.5307(0.1522) & \color{red} 0.0236(0.0594) & \color{red} 0.3345(0.2679) & \color{red} 0.0002(0.0010) &  \color{red} 0.0020(0.0451) \\ \hline
\multirow{2}{*}{Colon}
& \color{blue} 0.1316(0.0838) & \color{blue} \cellcolor{gray!15} 0.0384(0.0769) & \color{blue} 0.0701(0.0631) & \color{blue} \cellcolor{gray!40} 0.7995(0.0659) & \color{blue} \cellcolor{gray!40} 0.1465(0.0727) \\
& \color{red} 0.1298(0.0652) & \color{red} 0.0264(0.0519) & \color{red} 0.0685(0.0792) & \color{red} 0.0094(0.0188) & \color{red} 0.0000(0.0000) \\ \hline
\multirow{2}{*}{Bladder}
& \color{blue} 0.3960(0.1774) & \color{blue} 0.0972(0.1704) & \color{blue} 0.5234(0.2486) & \color{blue} \cellcolor{gray!40} 0.8230(0.1250) &  \color{blue} 0.0021(0.0451) \\
& \color{red} \cellcolor{gray!40} 0.5900(0.2343) & \color{red} 0.1016(0.1950) & \color{red} \cellcolor{gray!40} 0.6676(0.2430) & \color{red} 0.0001(0.0022) & \color{red} 0.0020(0.0451) \\ \hline
\multirow{2}{*}{Pancreas}
& \color{blue} 0.1233(0.0813) & \color{blue} \cellcolor{gray!40} 0.0743(0.1296) & \color{blue} \cellcolor{gray!40} 0.4007(0.1399) & \color{blue} \cellcolor{gray!40} 0.7512(0.0908) & \color{blue} 0.0115(0.0252) \\
& \color{red} \cellcolor{gray!40} 0.1589(0.0767) &  \color{red} 0.0299(0.0594) & \color{red} 0.0844(0.0894) & \color{red} 0.0001(0.0011) & \color{red} 0.0041(0.0638) \\ \hline
\multirow{2}{*}{Esophagus}
& \color{blue} \cellcolor{gray!40} 0.0904(0.0545) & \color{blue} \cellcolor{gray!15} 0.0356(0.0690) & \color{blue} \cellcolor{gray!40} 0.1721(0.0867) & \color{blue} \cellcolor{gray!40} 0.6927(0.0663) &  \color{blue} 0.0057(0.0472) \\
& \color{red} 0.0620(0.0321) & \color{red} 0.0235(0.0416) & \color{red} 0.0749(0.0662) & \color{red} 0.0000(0.0009) & \color{red} 0.0020(0.0451) \\ \hline
\multirow{2}{*}{Stomach}
& \color{blue} \cellcolor{gray!40} 0.4459(0.1482) & \color{blue} \cellcolor{gray!15} 0.0875(0.1680) & \color{blue} 0.3536(0.2567) & \color{blue} \cellcolor{gray!40} 0.8709(0.0986) & \color{blue} 0.0022(0.0036) \\
& \color{red} 0.1945(0.1477) & \color{red} 0.0546(0.1127) & \color{red} 0.3563(0.3034) & \color{red} 0.0023(0.0165) & \color{red} 0.0041(0.0638)  \\ \hline
\hline
\multirow{2}{*}{Avg}
& \color{blue} 0.4035(0.0417) & \color{blue} 0.1071(0.1376) & \color{blue} 0.4813(0.0737) & \color{blue} \cellcolor{gray!40} 0.8383(0.0532) & \color{blue} 0.0181(0.0207)\\
& \color{red} 0.4101(0.0516) & \color{red} 0.0747(0.0971) & \color{red} 0.3416(0.0999) & \color{red} 0.0103(0.0068) & \color{red} 0.0033(0.0440) \\
\hline \hline
	\end{tabular}
}
    \caption{Organ-wise Dice scores of the evaluated configurations in the PET/CT cohort.
    Results in blue and red are the performance of \textcolor{blue}{CT} and \textcolor{red}{PET}, respectively. Values are reported as mean (standard deviation).
    Light gray and dark gray shading denote the paired comparison markings at p$<$0.01 and p$<$0.0001, respectively.}
    \label{Table_PETCT}
\end{table*}
\begin{table*}[t]
	\centering
    \small
    \setlength\tabcolsep{6pt}
	\renewcommand\arraystretch{1.1}
    \resizebox{\textwidth}{!}{%
\begin{tabular}{c|ccccc}
		\hline 	\hline
Model & SAM-Med3D-turbo & SegVol & nnInteractive & VISTA3D & SAT  \\ \hline
\multirow{2}{*}{Liver}
& \color{teal} \cellcolor{gray!15} 0.7719(0.0755) & \color{teal} 0.3986(0.2433) & \color{teal} \cellcolor{gray!40} 0.8209(0.2412) & \color{teal} 0.0384(0.0548) & \color{teal} 0.0000(0.0005) \\
& \color{red} 0.6259(0.2050) & \color{red} 0.3731(0.2500) & \color{red} 0.1225(0.2252) & \color{red} \cellcolor{gray!40} 0.0749(0.0244) & \color{red} 0.0000(0.0000) \\ \hline
\multirow{2}{*}{Kidney-L}
& \color{teal} 0.4857(0.1422) & \color{teal} 0.1090(0.1136) & \color{teal} \cellcolor{gray!40} 0.6346(0.2242) & \color{teal} \cellcolor{gray!40} 0.0005(0.0053) & \color{teal} 0.0000(0.0000) \\
& \color{red} 0.5666(0.1142) & \color{red} 0.1047(0.0867) & \color{red} 0.1679(0.2398) & \color{red} 0.0000(0.0000) & \color{red} 0.0000(0.0000) \\ \hline
\multirow{2}{*}{Kidney-R}
& \color{teal} 0.5521(0.1488) & \color{teal} 0.0917(0.1088) & \color{teal} \cellcolor{gray!40} 0.7040(0.1546) & \color{teal} \cellcolor{gray!40} 0.0017(0.0090) & \color{teal} 0.0000(0.0000) \\
& \color{red} 0.5708(0.1257) & \color{red} 0.0884(0.0785) & \color{red} 0.2225(0.2494) & \color{red} 0.0000(0.0001) & \color{red} 0.0000(0.0000) \\ \hline
\multirow{2}{*}{Brain}
& \color{teal} 0.5797(0.1767) & \color{teal} 0.2342(0.1919) & \color{teal} 0.6405(0.3937) & \color{teal} 0.0000(0.0000) & \color{teal} \cellcolor{gray!40} 0.0378(0.0072) \\
& \color{red} 0.7707(0.0746) & \color{red} \cellcolor{gray!15} 0.2654(0.2306) & \color{red} \cellcolor{gray!40} 0.7860(0.3120) & \color{red} \cellcolor{gray!40} 0.0216(0.0180) & \color{red} 0.0000(0.0000) \\ \hline
\multirow{2}{*}{Heart}
& \color{teal} 0.6799(0.0893) & \color{teal} 0.0987(0.1069) & \color{teal} \cellcolor{gray!40} 0.5633(0.3423) & \color{teal} 0.0010(0.0040) & \color{teal} 0.0000(0.0000)  \\ 
& \color{red} 0.6685(0.1012) & \color{red} \cellcolor{gray!40} 0.2382(0.2124) & \color{red} 0.1430(0.2047) & \color{red} 0.0016(0.0066) & \color{red} 0.0000(0.0000)  \\ \hline
\multirow{2}{*}{Spleen}
& \color{teal} \cellcolor{gray!40} 0.5609(0.1130) & \color{teal} 0.1564(0.1477) & \color{teal} \cellcolor{gray!40} 0.6719(0.2582) & \color{teal} 0.0001(0.0017) & \color{teal} 0.0000(0.0000) \\
& \color{red} 0.4353(0.1606) & \color{red} 0.1655(0.1422) & \color{red} 0.1368(0.2115) & \color{red} 0.0004(0.0034) & \color{red} 0.0000(0.0000) \\ \hline
\multirow{2}{*}{Aorta}
& \color{teal} 0.0709(0.0458) & \color{teal} 0.0866(0.0872) & \color{teal} \cellcolor{gray!40} 0.2009(0.1384) & \color{teal} \cellcolor{gray!40} 0.0417(0.0681) & \color{teal} 0.0000(0.0000) \\
& \color{red} 0.0837(0.0437) & \color{red}  \cellcolor{gray!40} 0.1186(0.0811) & \color{red} 0.0465(0.0662) & \color{red} 0.0000(0.0000) & \color{red} 0.0000(0.0000) \\ \hline
\multirow{2}{*}{Lung}
& \color{teal} \cellcolor{gray!40} 0.6046(0.0938) & \color{teal} \cellcolor{gray!40} 0.2611(0.2298) & \color{teal} \cellcolor{gray!40} 0.5700(0.1768) & \color{teal} \cellcolor{gray!40} 0.0001(0.0006) & \color{teal} \cellcolor{gray!40} 0.0005(0.0022)  \\ 
& \color{red} 0.1892(0.1183) & \color{red} 0.0469(0.0718) & \color{red} 0.2685(0.2484) & \color{red} 0.0000(0.0001) & \color{red} 0.0000(0.0000) \\ \hline
\multirow{2}{*}{Colon}
& \color{teal} 0.1091(0.0586) & \color{teal} 0.0478(0.0648) & \color{teal} \cellcolor{gray!40} 0.1115(0.1110) & \color{teal} \cellcolor{gray!40} 0.1529(0.0785) & \color{teal} \cellcolor{gray!40} 0.0346(0.0556) \\
& \color{red} \cellcolor{gray!15} 0.1225(0.0661) & \color{red} 0.0416(0.0478) & \color{red} 0.0532(0.0632) & \color{red} 0.0029(0.0065) & \color{red} 0.0000(0.0000) \\ \hline
\multirow{2}{*}{Bladder}
& \color{teal} 0.6176(0.1912) & \color{teal} 0.0452(0.0863) & \color{teal} 0.5908(0.2813) & \color{teal} \cellcolor{gray!40} 0.0161(0.0617) & \color{teal} 0.0000(0.0000) \\
& \color{red} \cellcolor{gray!15} 0.6548(0.2074) & \color{red} \cellcolor{gray!40} 0.2613(0.2188) & \color{red} 0.5999(0.2067) & \color{red} 0.0000(0.0001) & \color{red} 0.0000(0.0000) \\ \hline
\multirow{2}{*}{Pancreas}
& \color{teal} 0.1642(0.0834) & \color{teal} 0.0579(0.0625) & \color{teal} \cellcolor{gray!40} 0.2961(0.1992) & \color{teal} \cellcolor{gray!40} 0.0301(0.0820) & \color{teal} 0.0043(0.0655) \\
& \color{red} 0.1702(0.0831) & \color{red} 0.0552(0.0559) & \color{red} 0.0554(0.0732) & \color{red} 0.0000(0.0000) & \color{red} 0.0043(0.0655) \\ \hline
\multirow{2}{*}{Esophagus}
& \color{teal} 0.0352(0.0209) & \color{teal} \cellcolor{gray!40} 0.0837(0.0678) & \color{teal} \cellcolor{gray!40} 0.1177(0.0936) & \color{teal} 0.0000(0.0000) & \color{teal} 0.0000(0.0000) \\
& \color{red} 0.0335(0.0169) & \color{red} 0.0435(0.0362) & \color{red} 0.0360(0.0448) & \color{red} 0.0000(0.0000) & \color{red} 0.0000(0.0000) \\ \hline
\multirow{2}{*}{Stomach}
& \color{teal} \cellcolor{gray!40} 0.4073(0.1306) & \color{teal} 0.0701(0.0925) & \color{teal} \cellcolor{gray!40} 0.7182(0.2040) & \color{teal} \cellcolor{gray!40} 0.0063(0.0218) & \color{teal} 0.0000(0.0000) \\
& \color{red} 0.2888(0.1211) & \color{red} \cellcolor{gray!40} 0.1204(0.0967) & \color{red} 0.1825(0.1667) & \color{red} 0.0005(0.0050) & \color{red} 0.0000(0.0000) \\ \hline
\hline
\multirow{2}{*}{Avg}
& \color{teal} 0.4334(0.0395) & \color{teal} 0.1340(0.0411) & \color{teal} 0.5107(0.0859) & \color{teal} \cellcolor{gray!40} 0.0222(0.0124) & \color{teal} 0.0059(0.0065) \\  
& \color{red} 0.3974(0.0552)  & \color{red} 0.1480(0.0427) & \color{red} 0.2170(0.0766) & \color{red} 0.0078(0.0027) & \color{red} 0.0003(0.0050)  \\
\hline \hline
	\end{tabular}
}
  \caption{Organ-wise Dice scores of the evaluated configurations in the PET/MRI cohort. Results in teal and red are the performance of \textcolor{blue}{MRI} and \textcolor{red}{PET}, respectively. Values are reported as mean (standard deviation). Light gray and dark gray shading denote the paired comparison markings at p$<$0.01 and p$<$0.0001, respectively.}
    \label{Table_PETMR}
\end{table*}

\subsection{Experimental settings}

We evaluated the official released weights and codebases for all evaluated foundation models including SAT \cite{SAT}, nnInteractive \cite{nninteractive}, SAM-Med3D \cite{SAM-Med3D}, VISTA3D \cite{vista3d} and SegVol \cite{segvol},  following the standardized evaluation protocols associated with each model. All experiments were conducted on an NVIDIA A100 GPU with 80GB of memory. 
We evaluate the zero-shot performance of these models without any fine-tuning, thereby assessing their out-of-the-box segmentation capability.
VISTA3D receives organ class-ID prompts and SAT receives organ names as text prompts. For the point-based systems, one positive point is sampled from the reference mask of each target.
We evaluate segmentation performance using Dice.
Implementation details are provided in the Appendix.

\section{Results}

\subsection{How does out-of-the-box performance vary across imaging modalities?}

Table~\ref{result_performance} shows substantial variation across imaging modalities and evaluated models. VISTA3D achieves the highest CT mean Dice among the foundation models (0.838), whereas nnInteractive achieves the highest MRI mean (0.511). SAM-Med3D-turbo has the highest PET means in both the PET/CT and PET/MRI cohorts (0.410 and 0.397, respectively). The highest-scoring configuration therefore changes with the imaging condition, highlighting the need to evaluate each tool on its intended input modality.

Strong structural imaging performance does not consistently extend to paired PET. VISTA3D scores 0.838 on CT but 0.010 on paired PET.  Similarly, nnInteractive scores 0.481 on CT and 0.342 on paired PET, and 0.511 on MRI and 0.217 on paired PET.  This direction is not universal: SAM-Med3D-turbo has similar CT and PET means in the PET/CT cohort, while SegVol scores slightly higher on PET than MRI in the PET/MRI cohort.  Nevertheless, PET scores remain limited overall, and some configurations also yield near-zero overlap on CT or MRI under the tested workflows.
In addition to foundation models, we developed task-specific nnUNets to provide a separate supervised reference. Trained with 10 labeled cases outside UMD, task-specific models achieves mean Dice scores of 0.686 on CT, 0.652 on MRI, and 0.605 and 0.560 on the corresponding PET cohorts. VISTA3D exceeds this reference on CT using its released pretrained weights, demonstrating that a foundation model can provide strong segmentation performance without task-specific training in a suitable imaging setting. On PET, however, all five evaluated foundation-model configurations score below the task-specific reference in both cohorts. 

In addition to the UMD evaluation, Table ~\ref{result_performance} includes test-set results reported in the original foundation model studies as contextual references. Although these evaluations differ in target organs and data distributions, the reported Dice scores summarized in the table all exceed 0.5.
Strong performance in the reported source-study evaluations does not consistently extend to the conditions evaluated on UMD, particularly when moving from structural to functional imaging.
Together with the modality discrepancy of training data composition in Fig. \ref{datafig}, the limited PET results indicate that generalization to underrepresented imaging modalities remains a challenge, where training coverage is one plausible contributing factor.

\subsection{Which organ-level capabilities are shared or different?}

Tables~\ref{Table_PETCT} and~\ref{Table_PETMR} show that some organ segmentation capabilities are maintained across structural and functional imaging, while others depend strongly on the modality and evaluated model.  SAM-Med3D-turbo achieves similar liver Dice scores on paired CT and PET (0.7039 and 0.6835) and similar heart scores on paired MRI and PET (0.6799 and 0.6685).  nnInteractive also produces relatively high brain segmentation scores across modalities: 0.7940 on CT and 0.8796 on paired PET, and 0.6405 on MRI and 0.7860 on paired PET. 
These examples show that the limitations observed on PET vary across organs.  One possible contributing factor is organ visibility: the brain’s high glucose metabolism and physiological FDG uptake produce a prominent PET signal, which may provide useful cues for segmentation.  This interpretation is consistent with the imaging characteristics described in the EANM brain PET guidelines \cite{guedj2022eanm}.

Other organ-level capabilities differ substantially between paired modalities, including within the same model.  In the PET/MRI cohort, nnInteractive scores 0.8209 on liver and 0.6719 on spleen using MRI, compared with 0.1225 and 0.1368 on paired PET.  This contrasts with its brain segmentation results in the same cohort, where PET scores exceed MRI scores.  Similarly, VISTA3D achieves a lung Dice of 0.9200 on CT but 0.0002 on paired PET under the tested workflow.  Thus, successful segmentation of an organ in structural imaging does not necessarily extend to its functional counterpart, and performance on one organ does not establish applicability to other targets.
The relative capabilities of different models can also change for the same organ across modalities. For the left kidney in the PET/CT cohort, nnInteractive scores higher than SAM-Med3D-turbo on CT (0.7930 versus 0.4701), whereas SAM-Med3D-turbo scores higher on paired PET (0.5678 versus 0.4464). A similar reversal occurs for the stomach in the PET/MRI cohort: nnInteractive has the higher MRI score, while SAM-Med3D-turbo has the higher PET score. 
The relative advantage of each model therefore depends on both the organ being segmented and the input modality.

Overall, some organ segmentation capabilities are maintained across structural and functional imaging, while others vary substantially with the modality and evaluated model.  By comparing corresponding organs within paired examinations, UMD helps distinguish these shared capabilities from those that remain specific to particular imaging conditions, providing a basis for further studies of organ-specific adaptation.

\begin{figure*}[t]
\centering
\includegraphics[width=\linewidth]{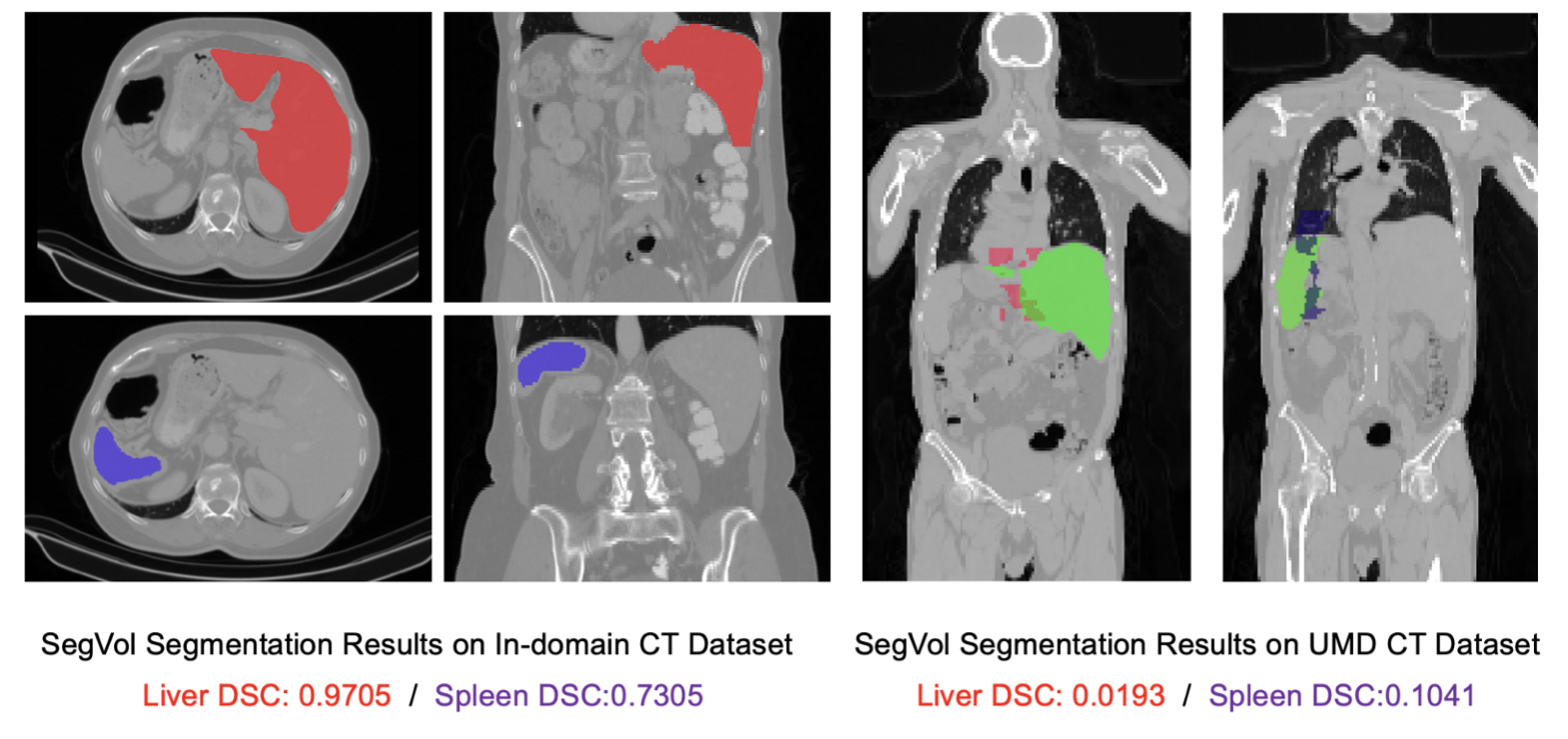}
\caption{Qualitative comparison of SegVol predictions on CT images from an imaging distribution
during training (left) and our UMD evaluation (right). The liver and spleen predictions illustrate differences in segmentation agreement across the two examples.}
\label{visualization}
\end{figure*}

\begin{figure*}[t]
\centering
\includegraphics[width=\linewidth]{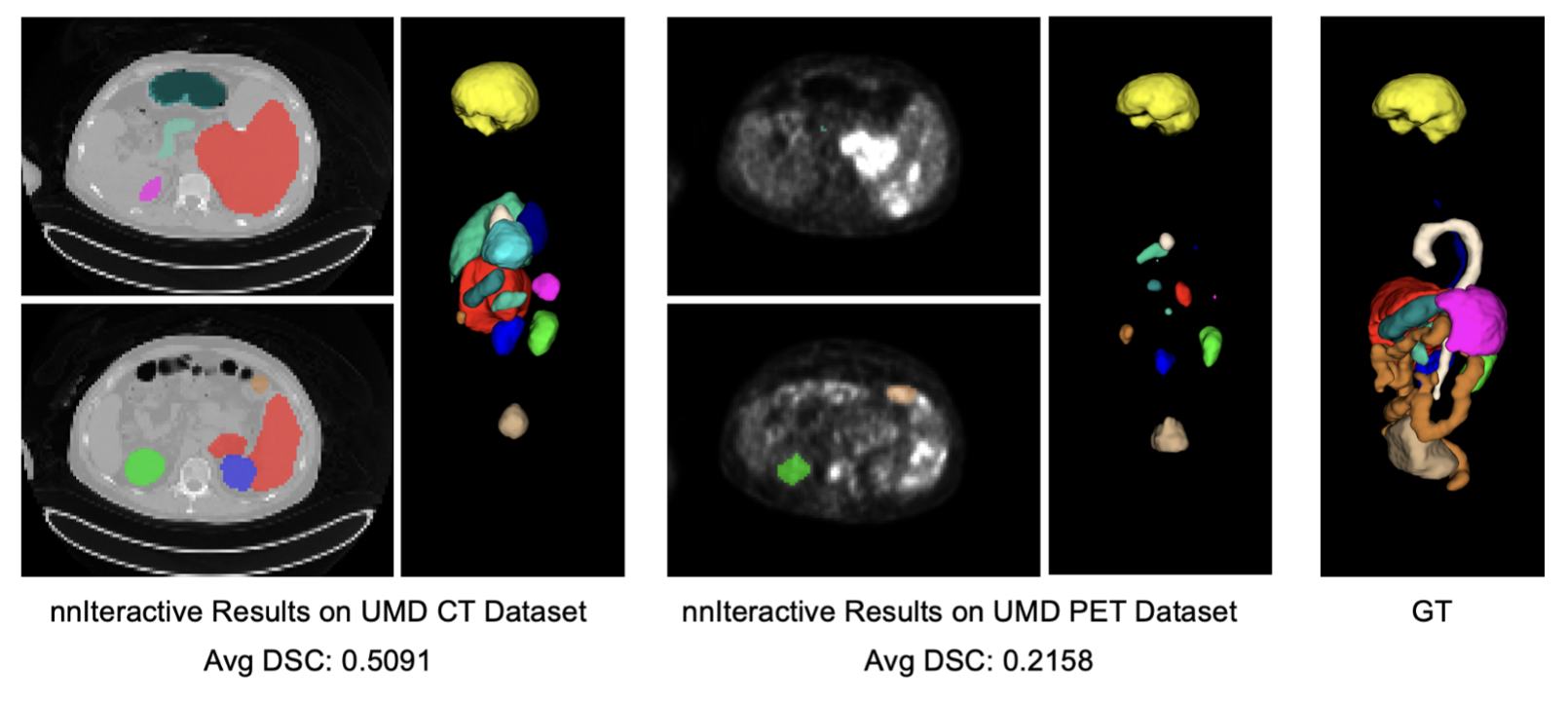}
\caption{
Qualitative comparison of nnInteractive predictions on paired CT and PET images from the same UMD examination. Two-dimensional overlays
and three-dimensional renderings illustrate differences in anatomical coverage, with the reference annotation shown on the right.}
\label{visualization2}
\end{figure*}

\subsection{Visualization of segmentation results across imaging conditions}

Following the modality-level and organ-level quantitative analysis, we present visualization of segmentation results across imaging conditions.
Figure~\ref{visualization} compares SegVol predictions on two CT examples.
The left example represents an imaging distribution encountered during SegVol training, where the liver and spleen predictions show good agreement with the reference annotations. In the illustrated UMD CT example, the predicted masks show limited overlap with the corresponding targets. These examples provide qualitative context for the variation
across datasets even when the imaging modality remains CT. 
The limited segmentation in this example is consistent with SegVol's low overall CT score on UMD.
Figure~\ref{visualization2} extends this comparison to structural and functional imaging. The CT and PET images come from the same examination and depict corresponding anatomical targets.
This example is consistent with the lower PET performance reported for nnInteractive. 
The visualizations show more complete anatomical coverage in the CT predictions and reduced coverage of several target organs in the PET predictions.

\begin{figure*}[t]
\centering
\includegraphics[width=\linewidth]{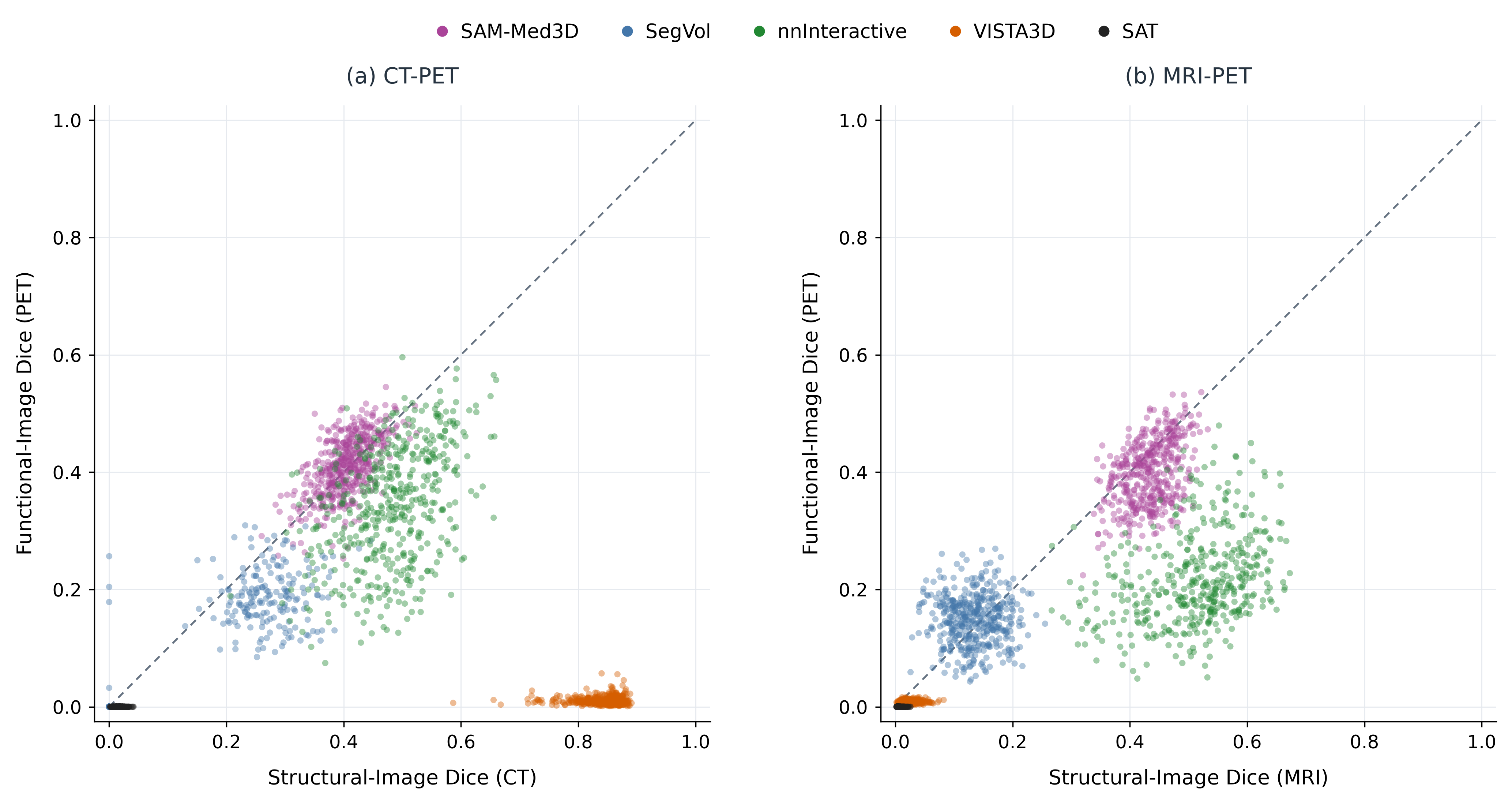}
\caption{Patient-level comparison of segmentation on structural and functional images. Each point represents one model evaluated on one examination, with Dice averaged over all evaluated organs on the structural image and paired PET.}
\label{scatter}
\end{figure*}

\subsection{Differences across individual examinations}

We compared structural-image and paired PET Dice at the examination level to assess whether the aggregate modality differences were confined to a few cases.   Using paired images and the same organ targets keeps patient identity and target coverage matched within each comparison.
As shown in Figure~\ref{scatter}, PET Dice was lower than CT Dice in 100\% of included examinations for VISTA3D and 94.8\% for nnInteractive. In the PET/MRI cohort, PET Dice was lower than MRI Dice in 99.5\%of examinations for nnInteractive and 72.3\% for SAM-Med3D. These proportions, together with the corresponding point distributions below the diagonal, indicate that lower PET performance in these configurations recurs across many examinations rather than being restricted to a few isolated observations. The direction and magnitude of the differences nevertheless varied by model and cohort. SAM-Med3D on PET/CT and SegVol on PET/MRI obtained higher PET scores in 57.2\% and 58.3\% of examinations,respectively. The predominant ordering therefore depends on the evaluated model and imaging condition.

\section{Discussion and Conclusion}

UMD provides a common evaluation setting for studying the applicability of segmentation tools across structural and functional imaging. The evaluation identifies useful out-of-the-box capabilities in selected structural-image settings, alongside substantial model- and organ-dependent limitations on paired PET. Patient-level comparisons further show that several modality differences recur across many examinations rather than being confined to isolated cases, although their direction and magnitude vary by model and cohort. These findings indicate that favorable performance in one imaging condition does not uniformly translate to another and that cross-modal applicability remains dependent on the evaluated task and workflow.
Future evaluations would extend the current zero-shot assessment to fine-tuning and other adaptation strategies. Shared patient splits and annotation budgets would enable comparisons of the benefits and costs of adapting foundation models to downstream segmentation tasks.
In addition, further work would include additional organs, acquisition settings, disease conditions, and tracers. Expanding evaluation beyond organ segmentation to tumors and other lesions would also allow cross-modal applicability to be examined for a broader range of targets.

In this work, our evaluation focuses on Dice as the primary measure of segmentation overlap, although it does not fully characterize boundary accuracy or the spatial distribution of errors \cite{maier2024metrics}. Most evaluated configurations obtained mean Dice below 0.6. Future benchmark tracks focusing on foundation-model adaptation will extend the evaluation beyond volumetric overlap. For adapted models achieving higher Dice scores, boundary-sensitive metrics will help quantify residual contour deviations and distinguish configurations with similar overlap but different boundary accuracy. This broader metric set will provide a more complete assessment of adaptation gains, including improvements that may be less apparent from Dice alone. 
Repeated prompt sampling and uncertainty estimates for paired comparisons could additionally characterize the stability of these gains across examinations and interaction choices. Documented configurations and standardized evaluation protocols will support comparable assessments across these extensions~\cite{bassi2024touchstone}.

In conclusion, UMD combines newly collected paired examinations, multi-organ annotations, and documented evaluation workflows to support systematic assessment of segmentation foundation models across structural and functional imaging.
By matching examinations and anatomical targets, it enables more direct comparisons of modality-associated performance differences and complements benchmarks in which modalities are evaluated on different tasks or cohorts. UMD provides a common setting for investigating cross-modal applicability and comparing future interaction and adaptation strategies under shared evaluation conditions.

\bibliographystyle{ieeenat_fullname}
\bibliography{main}

\begin{thebibliography}{23}
\providecommand{\natexlab}[1]{#1}
\providecommand{\url}[1]{\texttt{#1}}
\expandafter\ifx\csname urlstyle\endcsname\relax
  \providecommand{\doi}[1]{doi: #1}\else
  \providecommand{\doi}{doi: \begingroup \urlstyle{rm}\Url}\fi

\bibitem[Antonelli et~al.(2022)Antonelli, Reinke, Bakas, Farahani,
  Kopp-Schneider, Landman, Litjens, Menze, Ronneberger, Summers,
  et~al.]{antonelli2022medical}
Michela Antonelli, Annika Reinke, Spyridon Bakas, Keyvan Farahani, Annette
  Kopp-Schneider, Bennett~A Landman, Geert Litjens, Bjoern Menze, Olaf
  Ronneberger, Ronald~M Summers, et~al.
\newblock The medical segmentation decathlon.
\newblock \emph{Nature communications}, 13\penalty0 (1):\penalty0 4128, 2022.

\bibitem[Awais et~al.(2025)Awais, Naseer, Khan, Anwer, Cholakkal, Shah, Yang,
  and Khan]{awais2025foundation}
Muhammad Awais, Muzammal Naseer, Salman Khan, Rao~Muhammad Anwer, Hisham
  Cholakkal, Mubarak Shah, Ming-Hsuan Yang, and Fahad~Shahbaz Khan.
\newblock Foundation models defining a new era in vision: a survey and outlook.
\newblock \emph{IEEE Transactions on Pattern Analysis and Machine
  Intelligence}, 47\penalty0 (4):\penalty0 2245--2264, 2025.

\bibitem[Bassi et~al.(2024)Bassi, Li, Tang, Isensee, Wang, Chen, Chou, Roy,
  Kirchhoff, Rokuss, et~al.]{bassi2024touchstone}
Pedro~R Bassi, Wenxuan Li, Yucheng Tang, Fabian Isensee, Zifu Wang, Jieneng
  Chen, Yu-Cheng Chou, Saikat Roy, Yannick Kirchhoff, Maximilian Rokuss, et~al.
\newblock Touchstone benchmark: Are we on the right way for evaluating ai
  algorithms for medical segmentation?
\newblock \emph{Advances in Neural Information Processing Systems},
  37:\penalty0 15184--15201, 2024.

\bibitem[D'Antonoli et~al.(2024)D'Antonoli, Berger, Indrakanti, Vishwanathan,
  Weiß, Jung, Berkarda, Rau, Reisert, Küstner, Walter, Merkle, Segeroth,
  Cyriac, Yang, and Wasserthal]{totalsegmentator-mri}
Tugba~Akinci D'Antonoli, Lucas~K. Berger, Ashraya~K. Indrakanti, Nathan
  Vishwanathan, Jakob Weiß, Matthias Jung, Zeynep Berkarda, Alexander Rau,
  Marco Reisert, Thomas Küstner, Alexandra Walter, Elmar~M. Merkle, Martin
  Segeroth, Joshy Cyriac, Shan Yang, and Jakob Wasserthal.
\newblock Totalsegmentator mri: Sequence-independent segmentation of 59
  anatomical structures in mr images, 2024.

\bibitem[Du et~al.(2024)Du, Bai, Huang, and Zhao]{segvol}
Yuxin Du, Fan Bai, Tiejun Huang, and Bo Zhao.
\newblock Segvol: Universal and interactive volumetric medical image
  segmentation.
\newblock \emph{Advances in neural Information Processing Systems},
  37:\penalty0 110746--110783., 2024.

\bibitem[Gatidis et~al.(2024)Gatidis, Fr{\"u}h, Fabritius, Gu, Nikolaou,
  Foug{\`e}re, Ye, He, Peng, Bi, et~al.]{gatidis2024results}
Sergios Gatidis, Marcel Fr{\"u}h, Matthias~P Fabritius, Sijing Gu, Konstantin
  Nikolaou, Christian~La Foug{\`e}re, Jin Ye, Junjun He, Yige Peng, Lei Bi,
  et~al.
\newblock Results from the autopet challenge on fully automated lesion
  segmentation in oncologic pet/ct imaging.
\newblock \emph{Nature Machine Intelligence}, 6\penalty0 (11):\penalty0
  1396--1405, 2024.

\bibitem[Guedj et~al.(2022)Guedj, Varrone, Boellaard, Albert, Barthel, van
  Berckel, Brendel, Cecchin, Ekmekcioglu, Garibotto, et~al.]{guedj2022eanm}
Eric Guedj, Andrea Varrone, Ronald Boellaard, Nathalie~L Albert, Henryk
  Barthel, Bart van Berckel, Matthias Brendel, Diego Cecchin, Ozgul
  Ekmekcioglu, Valentina Garibotto, et~al.
\newblock Eanm procedure guidelines for brain pet imaging using [18f] fdg,
  version 3.
\newblock \emph{European journal of nuclear medicine and molecular imaging},
  49\penalty0 (2):\penalty0 632--651, 2022.

\bibitem[He et~al.(2025)He, Guo, Tang, Myronenko, Nath, Xu, Yang, Zhao, Simon,
  Belue, et~al.]{vista3d}
Yufan He, Pengfei Guo, Yucheng Tang, Andriy Myronenko, Vishwesh Nath, Ziyue Xu,
  Dong Yang, Can Zhao, Benjamin Simon, Mason Belue, et~al.
\newblock Vista3d: A unified segmentation foundation model for 3d medical
  imaging.
\newblock In \emph{Proceedings of the Computer Vision and Pattern Recognition
  Conference}, pages 20863--20873, 2025.

\bibitem[Ilesanmi et~al.(2024)Ilesanmi, Ilesanmi, and
  Ajayi]{ilesanmi2024reviewing}
Ademola~E Ilesanmi, Taiwo~O Ilesanmi, and Babatunde~O Ajayi.
\newblock Reviewing 3d convolutional neural network approaches for medical
  image segmentation.
\newblock \emph{Heliyon}, 10\penalty0 (6), 2024.

\bibitem[Isensee et~al.(2025)Isensee, Rokuss, Krämer, Dinkelacker, Ravindran,
  Stritzke, Hamm, Wald, Langenberg, Ulrich, et~al.]{nninteractive}
Fabian Isensee, Maximilian Rokuss, Lars Krämer, Stefan Dinkelacker, Ashis
  Ravindran, Florian Stritzke, Benjamin Hamm, Tassilo Wald, Moritz Langenberg,
  Constantin Ulrich, et~al.
\newblock nninteractive: Redefining 3d promptable segmentation.
\newblock \emph{arXiv preprint arXiv:2503.08373}, 2025.

\bibitem[Kirillov et~al.(2023)Kirillov, Mintun, Ravi, Mao, Rolland, Gustafson,
  Xiao, Whitehead, Berg, Lo, et~al.]{SAM}
Alexander Kirillov, Eric Mintun, Nikhila Ravi, Hanzi Mao, Chloe Rolland, Laura
  Gustafson, Tete Xiao, Spencer Whitehead, Alexander~C Berg, Wan-Yen Lo, et~al.
\newblock Segment anything.
\newblock In \emph{Proceedings of the IEEE/CVF international conference on
  computer vision}, pages 4015--4026, 2023.

\bibitem[Lalande et~al.(2022)Lalande, Chen, Pommier, Decourselle, Qayyum,
  Salomon, Ginhac, Skandarani, Boucher, Brahim, et~al.]{lalande2021deep}
Alain Lalande, Zhihao Chen, Thibaut Pommier, Thomas Decourselle, Abdul Qayyum,
  Michel Salomon, Dominique Ginhac, Youssef Skandarani, Arnaud Boucher, Khawla
  Brahim, et~al.
\newblock Deep learning methods for automatic evaluation of delayed
  enhancement-mri. the results of the emidec challenge.
\newblock \emph{Medical Image Analysis}, 79:\penalty0 102428, 2022.

\bibitem[Ma and Wang(2023)]{ma2023towards}
Jun Ma and Bo Wang.
\newblock Towards foundation models of biological image segmentation.
\newblock \emph{Nature Methods}, 20\penalty0 (7):\penalty0 953--955, 2023.

\bibitem[Ma et~al.(2022)Ma, Zhang, Gu, Zhu, Ge, Zhang, An, Wang, Wang, Liu,
  Cao, Zhang, Liu, Wang, Li, He, and Yang]{AbdomenCT-1K}
Jun Ma, Yao Zhang, Song Gu, Cheng Zhu, Cheng Ge, Yichi Zhang, Xingle An,
  Congcong Wang, Qiyuan Wang, Xin Liu, Shucheng Cao, Qi Zhang, Shangqing Liu,
  Yunpeng Wang, Yuhui Li, Jian He, and Xiaoping Yang.
\newblock Abdomenct-1k: Is abdominal organ segmentation a solved problem?
\newblock \emph{IEEE Transactions on Pattern Analysis and Machine
  Intelligence}, 44\penalty0 (10):\penalty0 6695--6714, 2022.

\bibitem[Maier-Hein et~al.(2024)Maier-Hein, Reinke, Godau, Tizabi, Buettner,
  Christodoulou, Glocker, Isensee, Kleesiek, Kozubek, et~al.]{maier2024metrics}
Lena Maier-Hein, Annika Reinke, Patrick Godau, Minu~D Tizabi, Florian Buettner,
  Evangelia Christodoulou, Ben Glocker, Fabian Isensee, Jens Kleesiek, Michal
  Kozubek, et~al.
\newblock Metrics reloaded: recommendations for image analysis validation.
\newblock \emph{Nature methods}, 21\penalty0 (2):\penalty0 195--212, 2024.

\bibitem[Mazurowski et~al.(2023)Mazurowski, Dong, Gu, Yang, Konz, and
  Zhang]{mazurowski2023segment}
Maciej~A Mazurowski, Haoyu Dong, Hanxue Gu, Jichen Yang, Nicholas Konz, and
  Yixin Zhang.
\newblock Segment anything model for medical image analysis: an experimental
  study.
\newblock \emph{Medical Image Analysis}, 89:\penalty0 102918, 2023.

\bibitem[Nioche et~al.(2018)Nioche, Orlhac, Boughdad, Reuz{\'e}, Goya-Outi,
  Robert, Pellot-Barakat, Soussan, Frouin, and Buvat]{nioche2018lifex}
Christophe Nioche, Fanny Orlhac, Sarah Boughdad, Sylvain Reuz{\'e}, Jessica
  Goya-Outi, Charlotte Robert, Claire Pellot-Barakat, Michael Soussan,
  Fr{\'e}d{\'e}rique Frouin, and Ir{\`e}ne Buvat.
\newblock Lifex: a freeware for radiomic feature calculation in multimodality
  imaging to accelerate advances in the characterization of tumor
  heterogeneity.
\newblock \emph{Cancer research}, 78\penalty0 (16):\penalty0 4786--4789, 2018.

\bibitem[Shen et~al.(2017)Shen, Wu, and Suk]{shen2017deep}
Dinggang Shen, Guorong Wu, and Heung-Il Suk.
\newblock Deep learning in medical image analysis.
\newblock \emph{Annual review of biomedical engineering}, 19:\penalty0
  221--248, 2017.

\bibitem[Wang et~al.(2025)Wang, Guo, Ye, Deng, Cheng, Li, Chen, Su, Huang,
  Shen, et~al.]{SAM-Med3D}
Haoyu Wang, Sizheng Guo, Jin Ye, Zhongying Deng, Junlong Cheng, Tianbin Li,
  Jianpin Chen, Yanzhou Su, Ziyan Huang, Yiqing Shen, et~al.
\newblock Sam-med3d: a vision foundation model for general-purpose segmentation
  on volumetric medical images.
\newblock \emph{IEEE Transactions on Neural Networks and Learning Systems},
  2025.

\bibitem[Wasserthal et~al.(2023)Wasserthal, Breit, Meyer, Pradella, Hinck,
  Sauter, Heye, Boll, Cyriac, Yang, Bach, and Segeroth]{totalsegmentator}
Jakob Wasserthal, Hanns-Christian Breit, Manfred~T. Meyer, Maurice Pradella,
  Daniel Hinck, Alexander~W. Sauter, Tobias Heye, Daniel~T. Boll, Joshy Cyriac,
  Shan Yang, Michael Bach, and Martin Segeroth.
\newblock Totalsegmentator: Robust segmentation of 104 anatomic structures in
  ct images.
\newblock \emph{Radiology: Artificial Intelligence}, 5\penalty0 (5):\penalty0
  e230024, 2023.

\bibitem[Zhang and Metaxas(2024)]{zhang2024challenges}
Shaoting Zhang and Dimitris Metaxas.
\newblock On the challenges and perspectives of foundation models for medical
  image analysis.
\newblock \emph{Medical image analysis}, 91:\penalty0 102996, 2024.

\bibitem[Zhang et~al.(2024)Zhang, Shen, and Jiao]{zhang2024segment}
Yichi Zhang, Zhenrong Shen, and Rushi Jiao.
\newblock Segment anything model for medical image segmentation: Current
  applications and future directions.
\newblock \emph{Computers in Biology and Medicine}, page 108238, 2024.

\bibitem[Zhao et~al.(2025)Zhao, Zhang, Wu, Zhang, Zhou, Zhang, Wang, and
  Xie]{SAT}
Ziheng Zhao, Yao Zhang, Chaoyi Wu, Xiaoman Zhang, Xiao Zhou, Ya Zhang, Yanfeng
  Wang, and Weidi Xie.
\newblock Large-vocabulary segmentation for medical images with text prompts.
\newblock \emph{NPJ Digital Medicine}, 8\penalty0 (1):\penalty0 566, 2025.

\end{thebibliography}

\section*{Implementation details}

All experiments were conducted on NVIDIA A100 GPUs with 80GB of memory. The released foundation models were evaluated without UMD-specific fine-tuning, using their respective prompting and inference workflows. Each imaging modality was supplied separately to the evaluated model. Here, zero-shot evaluation denotes the use of released pretrained weights without fine-tuning. Model-specific prompting and spatial preprocessing remain part of each evaluated workflow. The supervised nnUNet reference used additional training data outside UMD and was evaluated as a separate setting.

\noindent\textbf{SAM-Med3D-turbo.}
We evaluated the released turbo checkpoint of SAM-Med3D~\cite{SAM-Med3D} using the implementation at \url{https://github.com/uni-medical/SAM-Med3D}.
The inference scripts loaded \texttt{sam\_med3d\_turbo.pth} through the official implementation and its MedIM model loader.
Each image was reoriented to canonical orientation, resampled to $1.5$\,mm isotropic spacing, and z-score normalized over voxels with intensity greater than zero.
Reference masks were binarized for each target, and targets absent from the resampled reference mask were skipped.
Inputs were cropped or padded to $128\times128\times128$ voxels.
For each remaining target, one positive point was sampled from its reference mask; box and text prompts were disabled.
Predictions were thresholded at a sigmoid probability of 0.5, mapped back to the original image space, and saved as NIfTI masks with the input image shape, affine, and header.

\noindent\textbf{SegVol.}
We evaluated the point-based workflow of SegVol~\cite{segvol} using the implementation at \url{https://github.com/BAAI-DCAI/SegVol}.
The modality-specific inference scripts loaded \texttt{pytorch\_model.bin} through the SegVol implementation and its pretrained-model loader.
Clipping bounds and normalization statistics were computed from voxels above the image-wide mean.
Images were clipped to the 0.05th--99.95th percentiles of these voxels and standardized using their mean and standard deviation, followed by axis transposition, min--max scaling, and foreground cropping.
The global image and label views were resized to $32\times256\times256$ using nearest-neighbor interpolation.
Targets absent from the resized reference mask were skipped.
For each remaining target, one positive point was sampled from its reference mask, without additional positive or negative points; text and box prompts were disabled.
Zoom-out/zoom-in inference was enabled, with a local sliding-window overlap of 0.5.
Predictions were thresholded at a sigmoid probability of 0.5 and saved as NIfTI masks with the input image shape, affine, and header.
During evaluation, a missing prediction file was assigned a Dice score of zero when the corresponding reference target was nonempty.

\noindent\textbf{nnInteractive.}
We evaluated a single-positive-point workflow of nnInteractive~\cite{nninteractive} using the implementation at \url{https://github.com/MIC-DKFZ/nnInteractive}.
The modality-specific inference scripts initialized the nnInteractive v1.0 inference session using \texttt{checkpoint\_final.pth} from fold 0.
SimpleITK arrays were converted from $(z,y,x)$ to channel-first $(1,x,y,z)$ order.
The inference session cropped the nonzero image region and applied mean--standard-deviation normalization.
Automatic zoom was enabled, and \texttt{torch.compile} was disabled.
For each organ, interactions were reset, and one positive point was sampled uniformly from its reference mask using a NumPy random generator with seed 0.
Predicted masks were merged in ascending label order, with later masks overwriting overlapping regions, and saved with the input image geometry.

\noindent\textbf{VISTA3D.}
We evaluated the automatic class-ID interface of VISTA3D~\cite{vista3d} for the supported UMD organ targets using the implementation at \url{https://github.com/Project-MONAI/VISTA/tree/main/vista3d}.
The checkpoint \texttt{model.pt} was loaded using the MONAI bundle configuration \texttt{inference.json} and the corresponding modality-specific batch configuration.
The shared preprocessing configuration resampled images to $1.5$\,mm isotropic spacing, cropped the foreground with a 10-voxel margin, clipped intensities to $[-963.8248,1053.6785]$, linearly scaled them to $[0,1]$, and reoriented images to RAS.
Class-conditioned inference used the MONAI adaptive sliding-window inferer with $128\times128\times128$ windows, a sliding-window batch size of one, replicate padding, and automatic mixed precision.
The overlap was left at the inferer default.
The class-ID prompts were
$[1,14,5,22,115,3,6,28,29,30,31,32,62,15,4,11,12]$.
Predictions were mapped back to the input grid using nearest-neighbor interpolation.
Evaluation extracted the saved semantic class IDs, with lung labels 28--32 combined into one mask.

\noindent\textbf{SAT.}
We evaluated SAT~\cite{SAT} using the released SAT-Pro configuration and the implementation at \url{https://github.com/zhaoziheng/SAT}.
This configuration used the \texttt{UNET-L} visual backbone and knowledge-enhanced text encoder, with the released checkpoints \texttt{SAT\_Pro.pth} and \texttt{text\_encoder.pth}, respectively.
Each input was associated with a list of organ names and a modality identifier.
Organ names were converted to lowercase and encoded together with the modality information.
Images were reoriented to RAS, resampled to $1\times1\times3$\,mm spacing using linear interpolation, and cropped to the foreground.
CT intensities were clipped to $[-500,1000]$, whereas MRI and PET intensities were clipped to the 0.5th--99.5th percentiles of the cropped volume.
The clipped images were normalized to zero mean and unit standard deviation, and single-channel inputs were replicated across three channels.
Inference used $288\times288\times96$ patches with a nominal 50\% overlap; patches smaller than the inference window were zero-padded.
Inference used a patch batch size of two and a text-query batch size of up to 256.
Inference was performed under automatic mixed precision.
Patch-level sigmoid probabilities were combined by Gaussian-weighted averaging and thresholded at 0.5 to obtain per-organ binary masks.
The exporter saved these masks on the preprocessed image grid.

\end{document}